\documentclass{article}

\usepackage{arxiv}

\usepackage[utf8]{inputenc} 
\usepackage[utf8]{inputenc}   
\usepackage[T1]{fontenc}    
\usepackage{hyperref}       
\usepackage{url}            
\usepackage{booktabs}       
\usepackage{amsfonts}       
\usepackage{nicefrac}       
\usepackage{microtype}      
\usepackage{amsmath}
\usepackage{lipsum}
\usepackage{graphicx}
\usepackage{float}
\graphicspath{ {./images/} }

\title{Automated Extraction of Material Properties using LLM-based AI Agents}

\author{
  Subham Ghosh \\
  Mehta Family School of Data Science and Artificial Intelligence \\
  Indian Institute of Technology Roorkee \\
  Uttarakhand, India, 247667 \\
  \texttt{subham\_g1@mfs.iitr.ac.in} \\
  \And
  Abhishek Tewari \\
  Mehta Family School of Data Science and Artificial Intelligence \\
  Department of Metallurgical and Materials Engineering \\
  Indian Institute of Technology Roorkee \\
  Uttarakhand, India, 247667 \\
  \texttt{abhishek@mt.iitr.ac.in} \\
}

\begin{document}
\maketitle
\begin{abstract}
The rapid discovery of  materials is constrained by the lack of large, machine-readable datasets that couple performance metrics with structural context. Existing databases are either limited in scale, manually curated, or biased toward idealized first-principles results, leaving experimental literature underexploited. Here we present an \textbf{agentic, large language model (LLM) driven workflow} that autonomously extracts thermoelectric and structural properties from $\sim$10,000 full-text scientific articles. The system integrates dynamic token allocation, zero-shot multi-agent extraction, and conditional table parsing to balance accuracy against computational cost. Benchmarking on a manually curated set of 50 papers shows that GPT-4.1 achieves the highest extraction accuracy (F1 $\approx$ 0.91 for thermoelectric properties, F1 $\approx$ 0.82 for structural fields), while GPT-4.1 Mini offers nearly comparable performance at a fraction of the cost, enabling large-scale deployment. Applying this workflow, we create a data set of \textbf{ 27,822 property temperature records} with normalized units, spanning the figure of merit ($ZT$), Seebeck coefficient, conductivity, resistivity, power factor, and thermal conductivity, together with structural attributes such as crystal class, space group, and doping strategy. Analysis of the data set reproduces known thermoelectric trends, such as the superior performance of alloys over oxides and the advantage of p-type doping, while also surfacing broad structure property correlations. To facilitate community access, we release an \textbf{interactive web explorer} supporting semantic filters, numeric queries, and CSV export. Together, this study delivers the \textbf{largest LLM-curated thermoelectric dataset to date}, provides a reproducible and cost-profiled extraction pipeline, and establishes a foundation for scalable and data-driven thermoelectric material discovery. The workflow is broadly generalizable and can be adapted to curate structure–property datasets across diverse classes of functional materials beyond thermoelectrics.
\end{abstract}

\section{Introduction}

Materials informatics continues to face limited data readiness and accessibility. Although computational and experimental workflows can systematically generate new data, a vast body of historical results remains locked in published literature. As journal output grows, most information appears as unstructured prose and tables, impeding immediate reuse by modern, data-driven methods that require machine-readable, structured datasets.
In recent years, the application of natural language processing (NLP) and large language models (LLMs) to materials science has accelerated markedly. A central thrust has been the automated extraction of materials properties from the scientific literature, addressing a critical bottleneck in high-throughput materials discovery: data availability. Several domain-specific efforts have advanced materials text mining. MatBERT~\cite{trewartha2022quantifying} and ChemBERT~\cite{chithrananda2020chemberta} are domain-tuned transformers trained on large corpora of materials and chemistry papers, while MaterialsBERT~\cite{shetty2023general} integrates named-entity recognition (NER) layers to more precisely identify property terms.

Concurrently, general-purpose LLMs (e.g., GPT, Gemini, LLaMA) have demonstrated strong performance on classification, NER, and question-answering tasks even with limited training data owing to pretraining on broad scientific corpora that enables zero- and few-shot extraction. For example, Dagdelen \textit{et~al.} fine-tuned GPT-3.5 and LLaMA~2 to extract structured dopant–host relationships in MOFs~\cite{dagdelen2024structured}; Zheng \textit{et~al.} built a collaborative workflow with ChatGPT to extract over 26{,}000 synthesis parameters from 228 MOF papers~\cite{zheng2023chatgpt}; Polak and Morgan used prompt-chaining with GPT-4 to reduce hallucinations for metallic glasses and HEAs~\cite{polak2024extracting} , but their approach only operated on targeted sentences at a time, limiting the ability to capture cross-sentence relationships; Yang \textit{et~al.} repeatedly queried GPT-4 for band gaps, improving both accuracy and coverage over traditional datasets~\cite{yang2023accurate}; and Gupta \textit{et~al.} combined MaterialsBERT with GPT-3.5/LLaMA~2 to extract over one million polymer–property records from \(\sim\)681{,}000 full texts, explicitly evaluating accuracy, cost, and performance trade-offs~\cite{gupta2024data}. Ansari and Moosavi recently introduced Eunomia, a general agent-based LLM framework applied to case studies in MOFs and stability prediction~\cite{Ansari2024}. In Eunomia, the article text is tokenized and indexed with a vector database (FAISS), enabling retrieval of only the most relevant paragraphs for extraction. While Eunomia highlights the flexibility of multi-agent orchestration, it remains limited to text passages and small-scale demonstrations, without benchmarking cost–quality trade-offs or integrating tables and captions.

Thermoelectric discovery has traditionally relied on experiments and first-principles simulations (DFT, MD), which are accurate but slow and not readily scalable \cite{li2022machine,jain2016computational}. High-throughput screening helps \cite{gorai2017computationally,deng2024high,sarikurt2020high} yet remains costly for complex or doped systems \cite{chelikowsky2011computational}, motivating data-driven and ML approaches that leverage existing measurements \cite{jia2022unsupervised,wang2020machine,wang2023critical,antunes2023predicting,sparks2016data,mbaye2021data}. However, ML requires large, high-quality datasets: current public resources \cite{ricci2017ab,choudhary2020joint,yao2021materials,gorai2016te,wang2011assessing,carrete2014finding,xi2018discovery,fang2024wenzhou} skew toward ideal first-principles data, while experimental sets \cite{na2022public,katsura2019data,gaultois2013data,priya2021accelerated,lee2023texplorer} are small and manually curated. Most focus on a narrow subset of properties, underscoring the need for scalable, experimentally grounded datasets that couple multiple thermoelectric and structural attributes with consistent temperature context. Sierepeklis \textit{et~al.} used the rule-based ChemDataExtractor~\cite{mavracic2021chemdataextractor} to assemble 10{,}641 property records, highlighting challenges such as ambiguous units and composite descriptors~\cite{sierepeklis2022thermoelectric}. In thermoelectrics, by contrast, large-scale LLM-curated resources remain limited. More recently, Itani \textit{et~al.} employed GPTArticleExtractor to obtain 7{,}123 structured entries directly from full texts (e.g., ScienceDirect, Springer)~\cite{itani2025large,zhang2024gptarticleextractor}. While NER models reliably identify entities, they often struggle to capture cross-sentence relationships in scientific prose with complex, non-standard phrasing~\cite{olivetti2020data}. 

LLM-based extractors address some of these limitations but raise new challenges: inference is resource-intensive, requiring scalable pipelines that balance extraction quality against unnecessary calls. Existing systems further lack agentic controls such as candidate seeding, early exit, and dynamic token policies to manage reliability and cost, and few report transparent cost–quality benchmarks. Moreover, prior approaches have largely concentrated on narrative text alone, overlooking the rich quantitative data that frequently appears in tables and their captions. Together, these gaps motivate the development of an agentic, temperature-aware, and cost-profiled workflow that unifies text, tables, and captions for thermoelectric data curation at scale.

In this work, we develop an agentic LLM workflow purpose-built to extract thermoelectric properties together with their associated measurement temperatures and the structural descriptors of thermoelectric materials. Applied to \(\sim 10{,}000\) full-text articles, the pipeline curated a dataset of \textbf{27,822} records with normalized units at a total API cost of \(\mathbf{\$112}\). We benchmark multiple GPT and Gemini model families to quantify cost–quality trade-offs and select an operating point that balances accuracy and throughput. Using the curated corpus, we reveal dataset-level insights and release an open-source, web-based explorer that supports semantic and numeric-range filtering, row inspection, and CSV export. Together, these resources enable scalable structure–property analyses and provide a foundation for downstream machine-learning studies. This corpus constitutes, to our knowledge, the largest LLM-curated dataset of thermoelectric properties currently available. Beyond thermoelectrics, the workflow is readily generalizable to other materials domains by modifying prompt templates and property schemas.

\section{Methods}
\subsection{DOI Collection and Article Retrieval}

To build our dataset, we collected digital object identifiers (DOIs) for research articles related to thermoelectric materials. This was done by querying keywords such as “thermoelectric materials”, “ZT”, and “Seebeck coefficient”. We focused on three major scientific publishers: Elsevier, the Royal Society of Chemistry (RSC), and Springer.

After gathering the DOIs, we downloaded the corresponding articles using a combination of publisher APIs and web scraping techniques. Depending on availability, we retrieved either the xml or html version of each article. These structured formats were preferred in this work, as they are easier to process programmatically compared to pdf files~\cite{smith2022challenges}. Although recent end-to-end models like Nougat~\cite{blecher2023nougat} and Marker~\cite{paruchuri2023marker} have shown promising results in converting scientific pdfs into structured formats like Markdown, xml and html still offer more consistent parsing for large-scale automated extraction tasks. In total, we utilized approximately 10{,}000 open-access articles for this study.

\subsection{Preprocessing}

We developed an automated Python pipeline to preprocess scientific articles and extract key components such as full text, metadata, and tables\cite{oka2021machine} from both xml and html formats.

For Elsevier xml files, the pipeline uses structured xml tree traversal and regular expressions to accurately identify and extract table captions and rows. For html articles, a similar tag-based parsing approach is allowing the pipeline to handle varying layouts across publishers like Springer and RSC.

In the case of full text, we remove sections such as ``Conclusion'', ``References'', and other non-relevant portions that do not typically contain material property information. The remaining body text is then processed further to retain only the sentences that are likely to contain thermoelectric or structural properties.

This filtering is performed using a rule-based Python script which uses a large set of regular expression patterns. For making these expressions, we collected keywords related to material types (e.g. ``bulk'', ``nanoparticle''), thermoelectric properties (e.g., ``ZT'', ``Seebeck coefficient'', ``power factor''), structural parameters (e.g., ``lattice constants'', ``space group'', doping details), and common experimental methods. Then regular expression patterns were generated with assistance from  \texttt{ChatGPT}\cite{chatgpt2025} . Only sentences containing these patterns are retained and stored in a cleaned version of the article to ensure downstream LLM prompts are focused and token-efficient.

\begin{figure}[tbp]
    \centering 
    \includegraphics[width=1.0\textwidth]{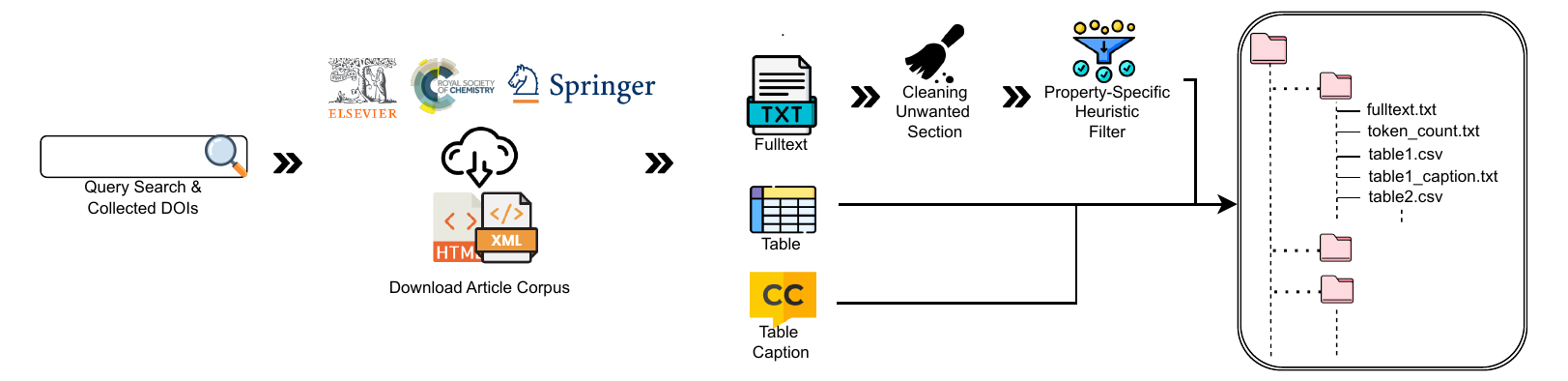} 
    \caption{Automated workflow for article retrieval and preprocessing, featuring document fetching, noise removal, tokenization, and metadata enrichment to generate LLM-ready datasets}
    \label{fig:preprocessing_workflow} 
\end{figure}

Finally, the script computes the number of tokens in the cleaned text using the \texttt{tiktoken}~\cite{tiktoken} tokenizer, and saves this count for downstream use in \texttt{max\_tokens}. Figure~\ref{fig:preprocessing_workflow} shows the hierarchical folder structure employed for data storage. This preprocessing strategy enables scalable, targeted, and efficient preparation of large corpora of materials science literature for property extraction tasks.

\subsection{The Data Extraction Workflow}

\begin{figure}[tbp]
    \centering 
    \includegraphics[width=1.0\textwidth]{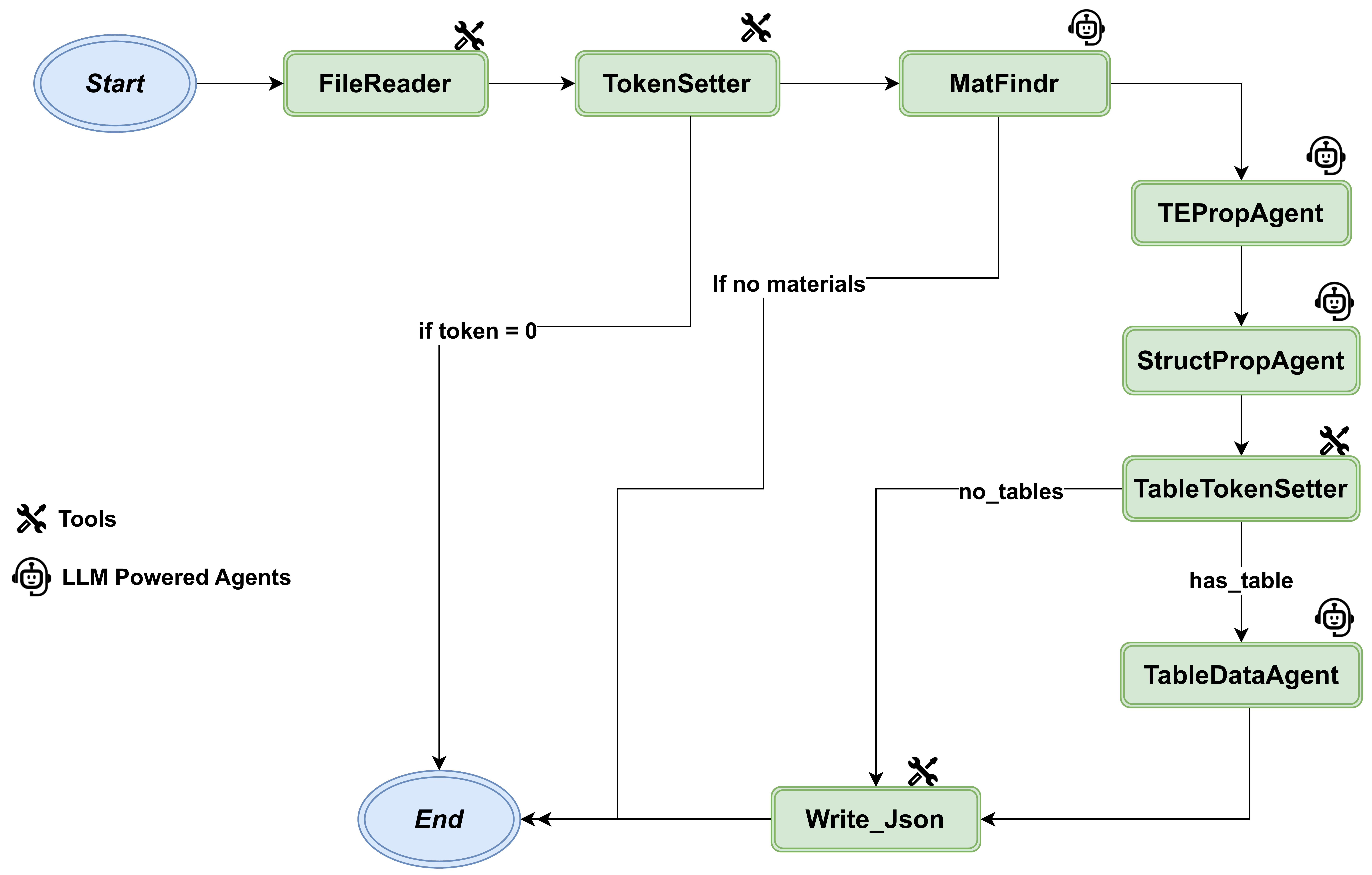} 
    \caption{Agentic LangGraph\cite{langgraph2024} workflow for extracting thermoelectric and structural properties using LLMs. The system dynamically allocates tokens, performs zero-shot extraction, and conditionally processes tabular data before saving structured outputs.}
    \label{fig:agentic_workflow} 
\end{figure}

Figure~\ref{fig:agentic_workflow} illustrates an agentic workflow based on the LangGraph\cite{langgraph2024} framework to extract thermoelectric and structural properties from full-text scientific articles. This process has four specialized LLM-based agents named as: material candidate finder (MatFindr) , thermoelectric property extractor (TEPropAgent), structural information extractor (StructPropAgent), and Table Data Extractor (TableDataAgent). Each agent fulfills a distinct role in parsing the article content, working in concert to reliably identify relevant information and populate our database with minimal human intervention. This modular “agent” design allows each step to focus on a well-defined sub-task, reducing complexity per query and enabling built-in checks and balances between stages. Notably, if an article describes multiple compounds, the workflow is capable of producing multiple structured entries in one pass – essentially a list of json objects, one for each material. The pipeline consists of multiple autonomous steps orchestrated as a state-based graph, allowing dynamic routing, conditional branching, and robust error handling during execution. In the following, we describe the function of each agent and the safeguards in place to ensure accurate and consistent data extraction.

The process begins with the ingestion of filtered full-text content from each article. A token analysis module computes the token count, which is then used to dynamically set the \texttt{max\_tokens} parameter for querying the language model (LLM) during data extraction. This enables efficient prompt sizing by balancing output completeness with API cost and latency.

The first agent MatFindr scans the full text of an article to identify all candidate thermoelectric materials mentioned. Its purpose is to build a definitive list of material names or formulas that will be the focus of subsequent extraction steps. We achieve this by prompting the LLM to recognize chemical formulas  and named compounds within the text, leveraging the model’s understanding of chemical nomenclature. To avoid spurious or trivial candidates, candidate validation is enforced by checking that each proposed material appears in context with relevant numerical data or units (e.g., the presence of a candidate’s name alongside terms like “Seebeck” or “ZT” in the text). This cross-check ensures that the list centers on materials for which thermoelectric data is actually reported, rather than every chemical mentioned. If no valid material is found, the agent signals an “early exit,” and the pipeline will skip further extraction for that article, thereby saving compute time and avoiding false entries.

For each material identified, the next agent, TEPropAgent, extracts its thermoelectric performance metrics from the article. This includes key properties such as Seebeck coefficient ($S$), electrical conductivity ($\sigma$), thermal conductivity ($\kappa$), power factor (PF), and the dimensionless figure of merit ($ZT$), as well as the temperatures at which these properties were measured (since thermoelectric properties are temperature-dependent). The extractor is implemented via a carefully engineered prompt that provides the LLM with two crucial inputs: (1) a focused context containing sentences or paragraphs where the given material is discussed, and (2) a structured template that explicitly lists the desired fields for output. By using the material’s name as a guiding hint in the prompt, we anchor the model’s attention to the correct entity and discourage it from drifting to unrelated content. After generation, a robust json parser monitors the outputs of the extractor agents: if the LLM’s response deviates from the expected json format (due to, say, an extra comma or a missing quote), the parser attempts to correct minor issues. 

A separate agent StructPropAgent focuses on structural attributes, including compound type, crystal structure, lattice parameters, space group, doping type, dopants, and processing method. By isolating this task, we prevent the intermixing of structural and performance information. The extractor also uses candidate material hints and robust json parser same as TEPropAgent. Then structural information is combined with the thermoelectric properties for each material, ready to be inserted into the database entry.

When tables are present , the workflow dynamically adjusts the token budget based on the number of table rows to ensure full coverage without exceeding context limits. Then incorporates TableDataAgent that specifically handles the content of tables. Tables and captions are reformatted into a structured text representation before prompting the model. It also uses candidate material hints and robust json parser as before. The TableDataAgent extracts both thermoelectric and structural data from the tabular content, returning results in the same schema, that can be directly compared or merged with the results from the text-based extractors. As a further check, the workflow compares the data obtained from tables with any values extracted from the main text. In many cases, tables serve to summarize data also described in the text; when both sources are available, we verify that they are consistent. Our approach favors information explicitly stated in the text (since context clarifies its meaning), but if a property is only found in a table and not in the narrative, the table extractor’s result is used to fill that gap.

The entire pipeline is implemented as a LangGraph~\cite{langgraph2024} state machine, where each node corresponds to a functional module (e.g., read, extract, route, write), and conditional transitions support dynamic decisions, such as skipping table extraction when no tables are present. All outputs—including thermoelectric properties, structural parameters, and table-derived data—are stored in structured json format for downstream use. 

For each agent, we craft task-specific prompts and issue them in a zero-shot manner, providing the extracted material names as explicit context to focus the model’s attention and reduce irrelevant outputs. We set the temperature to $T=0.001$ to minimize stochasticity and enforce deterministic responses. This approach aligns with prior studies showing that lower temperatures produce more focused, reproducible outputs suited for precise information extraction tasks~\cite{renze2024effect,zhu2024hot}. Employ a dynamic token allocation strategy, where the \texttt{max\_tokens} parameter is adapted based on the length of the input. This ensures efficient utilization of the LLM context window, balancing cost and latency with output completeness, as suggested in prior work on adaptive prompting and efficient large-context usage~\cite{liu2023lost}. 

Although developed for thermoelectric materials, the workflow is inherently extensible. By modifying prompt templates and property schemas, the same architecture can be applied to other scientific domains. Zero-shot prompting enables rapid adaptation of the workflow to new domains. For example, in catalysis one could extract reaction energies and turnover frequencies, or in battery research one could target capacity, cycle life, and Coulombic efficiency.  This modular agentic, LLM-powered workflow demonstrates multi-step reasoning, autonomous decision-making, adaptive token control, and modular API-based interaction—making it a scalable and extensible foundation for high-throughput scientific information extraction.
\section{Performance Evaluation and Model Comparison}

We evaluate the reliability and efficiency of our agentic extraction pipeline on a manually curated benchmark of \textbf{50 full-text thermoelectric articles}, covering both \textbf{thermoelectric (TE) properties} (numerical, tolerance-based matching) and \textbf{structural descriptors} (categorical, ontology-guided matching). Results are reported for four state-of-the-art LLMs: \textbf{GPT-4.1}, \textbf{GPT-4.1 Mini}, \textbf{Gemini 1.5 Pro}, and \textbf{Gemini 2.0 Flash}.

\subsection{TE Property Evaluation}

For TE properties, we benchmark the three most frequently reported metrics: figure of merit (\textbf{ZT}), \textbf{Seebeck coefficient (S)}, and \textbf{thermal conductivity} ($\kappa$). Each property is evaluated together with its measurement temperature when available.

\paragraph{Convergence criterion.}
A predicted value \(p\) at temperature \(t_p\) is considered a \textbf{True Positive (TP)} if it matches any \emph{unmatched} ground-truth value \(g\) at temperature \(t_g\) such that
\begin{equation}
\frac{|g - p|}{\max(|g|, |p|, 10^{-6})} \leq 0.01
\quad \text{and} \quad
\left( \ |t_p - t_g| \leq 1\,\mathrm{K} \right).
\label{eq:convergenc_criteria}
\end{equation}
If no such ground truth exists for a given prediction (considering both value and temperature constraints), it is counted as a \textbf{False Positive (FP)}. Conversely, any unmatched ground-truth values are counted as \textbf{False Negatives (FN)}. \textit{If either \(t_g\) or \(t_p\) is unavailable}, we apply only the value-tolerance term in Eq.~\eqref{eq:convergenc_criteria}.

\paragraph{Metrics.}
Using these definitions, we compute:
\begin{align}
\text{Precision} &= \frac{\text{TP}}{\text{TP} + \text{FP}}, \\
\text{Recall} &= \frac{\text{TP}}{\text{TP} + \text{FN}}, \\
\text{F1-score} &= \frac{2 \cdot \text{Precision} \cdot \text{Recall}}{\text{Precision} + \text{Recall}}.
\end{align}

\begin{table}[tbp]
\centering
\begin{tabular}{|l|c|c|c|c|}
\hline
\textbf{Model} & \textbf{ZT} & \textbf{Seebeck (S)} & \textbf{Thermal Conductivity ($\kappa$)} & \textbf{Overall (Micro)} \\
\hline
{Gemini 1.5 Pro} 
& P = 0.924 & P = 0.925 & P = 0.902 & P = 0.918\\
& R = 0.802 & R = 0.661 & R = 0.868 & R = 0.780 \\
& F1 = 0.859 & F1 = 0.771 & F1 = 0.885 & F1 = 0.843 \\
\hline
{Gemini 2.0 Flash} 
& P = 0.897 & P = 0.977 & P = 0.837 & P = 0.903\\
& R = 0.670 & R = 0.750 & R = 0.679 & R = 0.695\\
& F1 = 0.767 & F1 = 0.848 & F1 = 0.750 & F1 = 0.785 \\
\hline
{GPT-4.1 Mini} 
& P = 0.885 & P = 0.925 & P = 0.879 & P = 0.894\\
& R = 0.846 & R = 0.875 & R = 0.962 & R = 0.885\\
& F1 = 0.865 & F1 = 0.899 & F1 = 0.919 & F1 = 0.889 \\
\hline
{GPT-4.1} 
& \textbf{P = 0.909} & \textbf{P = 0.961} & \textbf{P = 0.895} & \textbf{P = 0.918}\\
& \textbf{R = 0.879} & \textbf{R = 0.875} & \textbf{R = 0.962} & \textbf{R = 0.900} \\
& \textbf{F1 = 0.894} & \textbf{F1 = 0.916} & \textbf{F1 = 0.927} & \textbf{F1 = 0.909} \\
\hline
\end{tabular}
\vspace{0.5em}
\caption{Performance comparison on the \textbf{thermoelectric property extraction} task (50 papers). Each entry reports Precision (P), Recall (R), and F1-score (F1) for the respective property. The \textbf{Overall} column reports micro-averaged metrics across TE properties.}
\label{tab:precision_recall}
\end{table}

\noindent
The results in Table~\ref{tab:precision_recall} reveal clear differences in
model behavior across thermoelectric properties. GPT-4.1 consistently delivers the strongest
overall performance, with high precision and recall across all three properties and a balanced
F1-score of 0.909. GPT-4.1 Mini performs nearly as well (F1 = 0.889), demonstrating only a
marginal drop in accuracy despite its smaller size and substantially lower cost. This indicates
that for most large-scale extraction tasks, GPT-4.1 Mini offers an attractive balance of accuracy
and efficiency.

In contrast, both Gemini models exhibit weaker recall, particularly for Seebeck coefficient
extraction (R = 0.661 for Gemini 1.5 Pro and R = 0.750 for Gemini 2.0 Flash). This suggests
that Gemini models are more conservative, often missing valid ground-truth values even when
precision remains high. Thermal conductivity is the most robustly extracted property across all
models (F1 $\approx$ 0.88–0.93), while Seebeck coefficient shows the greatest variability,
likely due to broader linguistic diversity in its reporting. ZT extraction lies in between, with
performance strongly correlated to how explicitly ZT values are stated in text or tables.
Overall, GPT-4.1 sets the benchmark for reliability, while GPT-4.1 Mini offers competitive
performance at lower cost, and Gemini models show uneven extraction quality across properties.

\subsection{Structural Property Evaluation}

Structural descriptors (\textit{lattice structure}, \textit{compound type}, and \textit{doping type}) were evaluated using a tailored hybrid benchmarking framework that accounts for the linguistic diversity of categorical fields. For \textit{lattice structure} and \textit{compound type}, we combined ontology-based normalization with semantic embeddings (all-MiniLM-L6-v2) and a logistic regression classifier. This setup ensures that near-synonymous expressions are resolved correctly, for example, ``rocksalt,'' ``rock-salt structure,'' and ``face-centered cubic'' are consistently mapped to the canonical \textit{fcc} class, while ``layered perovskite'' and ``Ruddlesden–Popper'' are aligned under the perovskite family. For \textit{doping type}, where interpretation depends on identifying specific dopant elements and their electronic role (donor vs.\ acceptor), we implemented rule-based heuristics using a curated dopant dictionary. This allows us to capture cases such as ``La-doped BaTiO$_3$'' (classified as n-type), ``Na-doped PbTe'' (p-type), and more complex examples like ``co-doped with Li and Nb,'' which are treated as compensated systems. To avoid penalizing superficial differences, we further applied relaxed equivalences (e.g., ``p'' $\approx$ ``p-type,'' ``n'' $\approx$ ``n-type''). Together, this hybrid strategy of ontologies, semantic embeddings, and domain-specific rules provides both precision and flexibility in benchmarking structural extractions. The extraction quality for structural fields was quantified using the same metrics as for thermoelectric properties, namely \textbf{Precision (P)}, \textbf{Recall (R)}, and \textbf{F1-score (F1)} against the manually curated ground truth.

\begin{table}[tbp]
\centering
\begin{tabular}{|l|c|c|c|c|}
\hline
\textbf{Model} & \textbf{Lattice Structure} & \textbf{Compound Type} & \textbf{Doping Type} & \textbf{Overall (Macro)} \\
\hline
{Gemini 1.5 Pro} 
& P = 0.882 & P = 0.795 & P = 0.506 & P = 0.728 \\
& R = 0.882 & R = 0.795 & R = 0.506 & R = 0.728 \\
& F1 = 0.882 & F1 = 0.795 & F1 = 0.506 & F1 = 0.728 \\
\hline
{Gemini 2.0 Flash} 
& P = 0.889 & P = 0.707 & P = 0.628 & P = 0.741 \\
& R = 0.889 & R = 0.707 & R = 0.628 & R = 0.741 \\
& F1 = 0.889 & F1 = 0.707 & F1 = 0.628 & F1 = 0.741 \\
\hline
{GPT-4.1 Mini} 
& P = 0.938 & P = 0.925 & P = 0.562 & P = 0.808 \\
& R = 0.938 & R = 0.925 & R = 0.562 & R = 0.808 \\
& F1 = 0.938 & F1 = 0.925 & F1 = 0.562 & F1 = 0.808 \\
\hline
{GPT-4.1} 
& \textbf{P = 0.931} & \textbf{P = 0.880} & \textbf{P = 0.639} & \textbf{P = 0.817} \\
& \textbf{R = 0.931} & \textbf{R = 0.880} & \textbf{R = 0.639} & \textbf{R = 0.817} \\
& \textbf{F1 = 0.931} & \textbf{F1 = 0.880} & \textbf{F1 = 0.639} & \textbf{F1 = 0.817} \\
\hline
\end{tabular}
\caption{Performance comparison on the \textbf{structural property extraction} task (50 papers). Each entry reports P/R/F1 for the respective field. \textbf{Overall} values are macro-averaged across structural fields.}
\label{tab:structural}
\end{table}

\noindent
Table~\ref{tab:structural} highlights distinct trends across structural fields.
For lattice structure, all models achieve high and consistent performance (F1 $\approx$ 0.88–0.94),
with GPT-4.1 Mini slightly outperforming GPT-4.1 (0.938 vs.\ 0.931). This demonstrates that
ontology-guided embeddings and classifier support are sufficient to resolve common synonyms
and variants (e.g., ``rocksalt,'' ``rock-salt structure,'' ``fcc'') across models.

Compound type extraction also performs strongly, with GPT-4.1 Mini again achieving the
highest F1 (0.925). Gemini models lag behind (0.707–0.795), reflecting their difficulty in
disambiguating overlapping chemical families such as ``semiconductor'' versus ``alloy.'' This
suggests that compound type requires both broad chemical context and nuanced classification
capability, favoring GPT-based models.

The most challenging field is doping type, where all models show reduced performance
(F1 = 0.51–0.64). GPT-4.1 achieves the best score (0.639), but the gap compared to lattice
and compound type indicates that even advanced models struggle to incorporate chemical
knowledge about dopants (e.g., La $\rightarrow$ n-type, Na $\rightarrow$ p-type). Rule-based
heuristics mitigate this to some extent, but co-doping and compensation cases remain
problematic. Here, Gemini 1.5 Pro performs particularly poorly (0.506), reflecting both
lower recall and weaker handling of implicit information.

Across fields, GPT-4.1 provides the most balanced accuracy, while GPT-4.1 Mini
consistently matches or exceeds it for lattice and compound classification. The Gemini
models, by contrast, underperform across categorical descriptors, with especially low
accuracy for compound and doping type. These results emphasize the complementary
strengths of GPT models for structural benchmarking, while underscoring the need for
deeper domain-knowledge integration in doping-type classification.

Taken together, these results highlight that GPT-4.1 is the optimal choice for targeted 
applications where maximal accuracy is essential (e.g., constructing high-quality benchmarks 
or validating edge cases). However, for large-scale corpus extraction, \textbf{GPT-4.1 Mini 
emerges as the most efficient operating point}, providing nearly the same performance 
as GPT-4.1 while reducing API costs by a factor of 5--10. This balance of scalability, 
affordability, and robustness motivates our decision to adopt GPT-4.1 Mini for dataset 
construction, while reserving GPT-4.1 for precision-critical studies.

Taken together, the benchmarking results reveal a clear cost--quality gradient across model families. 
For thermoelectric properties, GPT-4.1 achieves the highest overall accuracy with $F_{1}=0.909$, 
while GPT-4.1 Mini follows closely at $F_{1}=0.889$, representing only a $\sim$2\% drop despite 
its much smaller size. For structural descriptors, GPT-4.1 again leads with $F_{1}=0.817$, with 
GPT-4.1 Mini nearly matching it at $F_{1}=0.808$. In contrast, Gemini 2.0 Flash and Gemini 1.5 Pro 
perform less consistently, trailing behind at $F_{1}=0.785/0.741$ and $F_{1}=0.843/0.728$ for 
thermoelectric/structural tasks, respectively. This trend shows that GPT-based models provide 
both higher recall and more balanced accuracy across numerical and categorical fields.
\begin{figure}[tbp]
    \centering
    \includegraphics[width=0.8\linewidth]{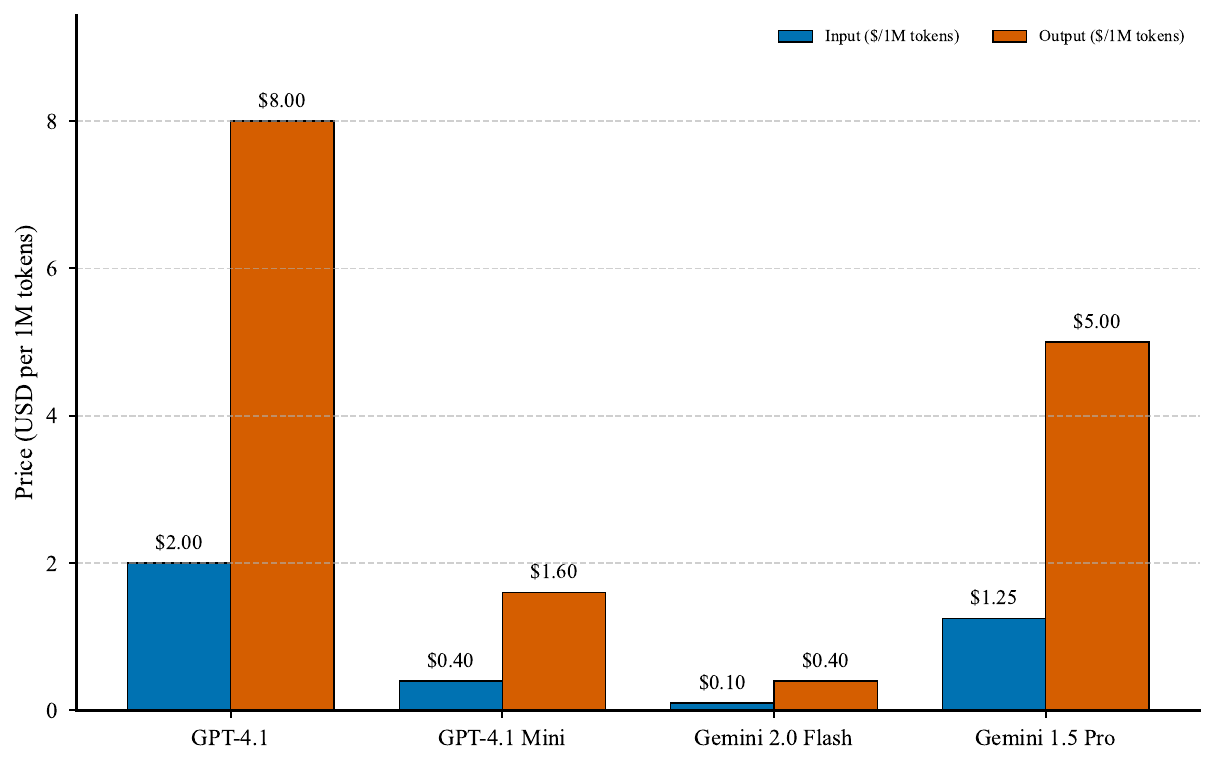}
    \caption{Token pricing comparison for \textbf{GPT}\cite{openai_pricing} and \textbf{Gemini}\cite{google_gemini_pricing} models (input/output cost per 1M tokens).}
    \label{fig:llm_pricing}
\end{figure}
The pricing comparison in Fig.~\ref{fig:llm_pricing} underscores these trade-offs. Running 
GPT-4.1 at scale incurs nearly an order of magnitude higher cost per million tokens compared 
to GPT-4.1 Mini, translating to a practical difference of hundreds of dollars when processing 
$\sim$10,000 full-text articles. Given the marginal accuracy gap (0.909 vs.\ 0.889 for TE; 0.817 
vs.\ 0.808 for structural), GPT-4.1 Mini offers a far more favorable balance of cost and 
performance, making it the pragmatic choice for large-scale dataset construction. GPT-4.1 
remains best reserved for precision-critical benchmarks or cases where maximal recall is essential.

This cost-quality analysis motivates our use of GPT-4.1 Mini for corpus-scale extraction, while 
recognizing GPT-4.1 as the highest-accuracy model. In the following sections, we demonstrate how 
this choice enables the creation of the largest LLM-curated thermoelectric dataset to date, while 
maintaining both affordability and robustness.

\section{Dataset Curation and Analysis}

We curated a large-scale thermoelectric dataset from nearly 10,000 scientific articles to demonstrate the performance of the agentic workflow described above with \textbf{GPT-4.1 Mini} for text and table extraction. Each article was parsed to generate structured records of thermoelectric (TE) and structural properties. The postprocessing steps involved the removal of spurious fields occasionally introduced by LLMs and the retention of only entries containing at least one TE property. Each \texttt{doi} and its corresponding materials were considered as dataset keys. The final dataset covers about 27822 rows.

\begin{figure}[tbp]
    \centering
    \includegraphics[width=0.95\textwidth]{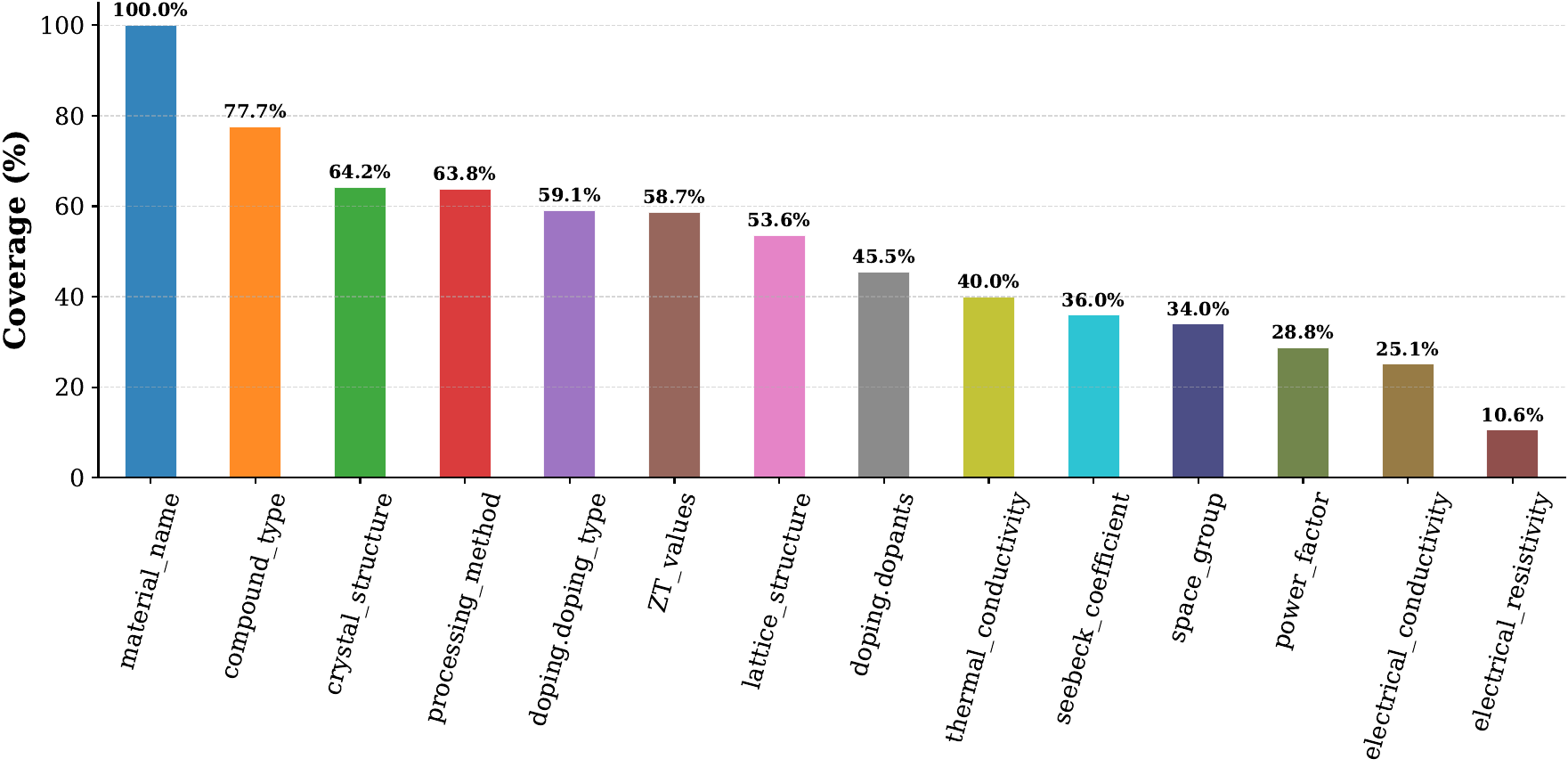}
    \caption{Coverage percentage of extracted thermoelectric and structural properties across the curated dataset.}
    \label{fig:coverage}
\end{figure}

The dataset includes the primary TE properties: Figure of merit($ZT$), Seebeck coefficient ($S$), electrical conductivity ($\sigma$), electrical resistivity ($\rho$), power factor (PF), and thermal conductivity ($\kappa$), with corresponding temperatures wherever available. Since $\sigma$ and $\rho$ are inversely related, they can be treated as one combined property. The structural domain encompasses compound type, crystal structure, lattice structure, space group, processing method, doping type, and dopants. This joint representation of thermoelectric performance with structural and doping information enables comprehensive structure–property correlations. Figure~\ref{fig:coverage} shows the coverage percentage for each property. The relatively higher coverage of ZT values in comparison to other thermoelectric properties could be due to general trend of ZT values being discussed in the text of the papers more frequently, while other properties are usually represented in the figures. The analysis also shows a nearly uniform coverage of the important structural attributes reported in the papers. 

\subsection{Thermoelectric Properties}
\begin{figure}[tbp]
    \centering
    \includegraphics[width=0.95\textwidth]{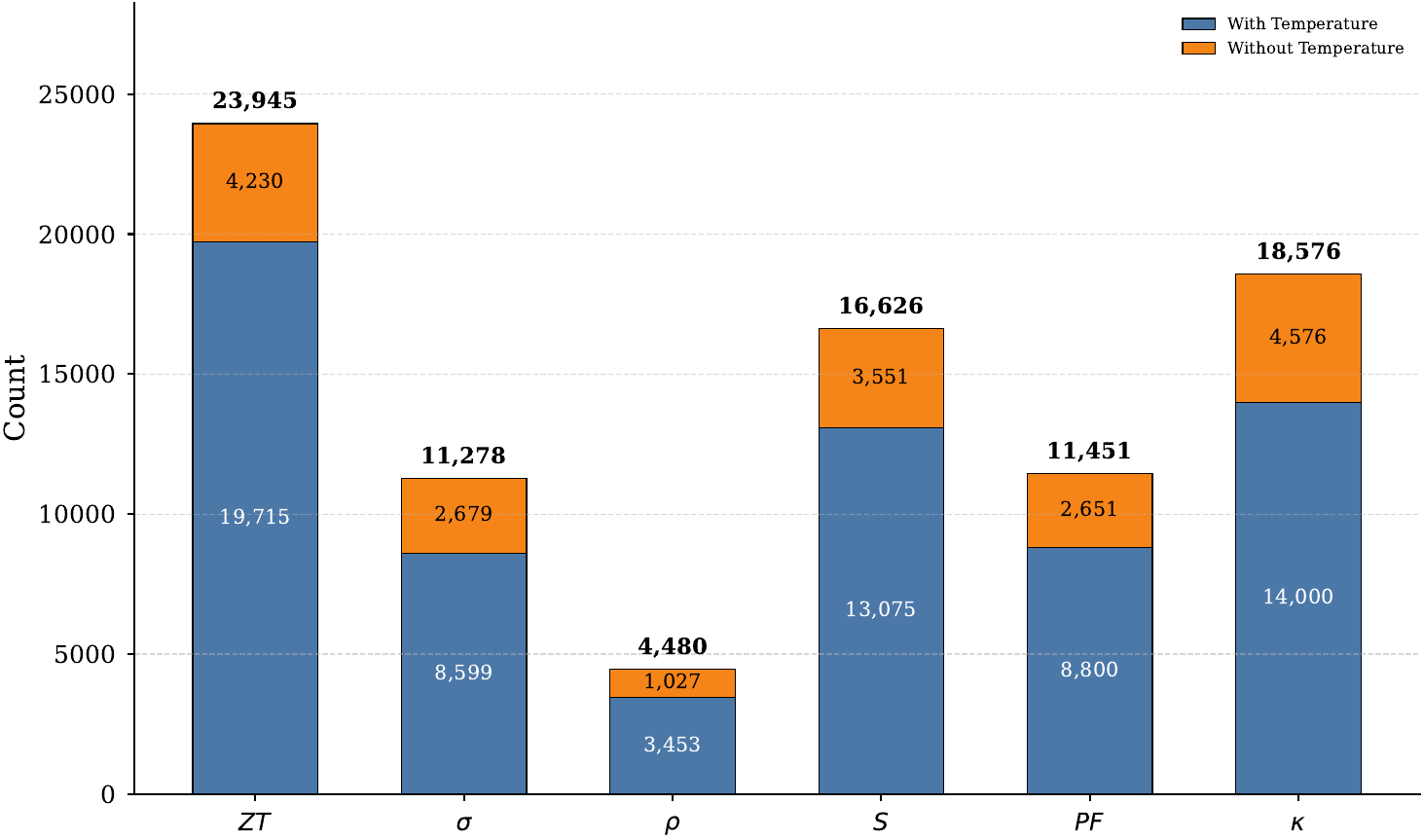}
    \caption{Counts of thermoelectric property records with and without corresponding temperature information.}
    \label{fig:TE_prop&temp}
\end{figure}

As all thermoelectric properties are inherently temperature dependent, recording the corresponding measurement temperature is essential. For each material, we retain all available property-temperature pairs when multiple measurements are reported. This ensures that the dataset not only captures the property values but also their thermal context, enabling a more accurate comparison across materials and conditions. Figure~\ref{fig:TE_prop&temp} presents the distribution of property records with and without associated temperatures, showing that a majority of entries include temperature annotations.  

\begin{figure}[tbp]
    \centering
    \includegraphics[width=0.95\textwidth]{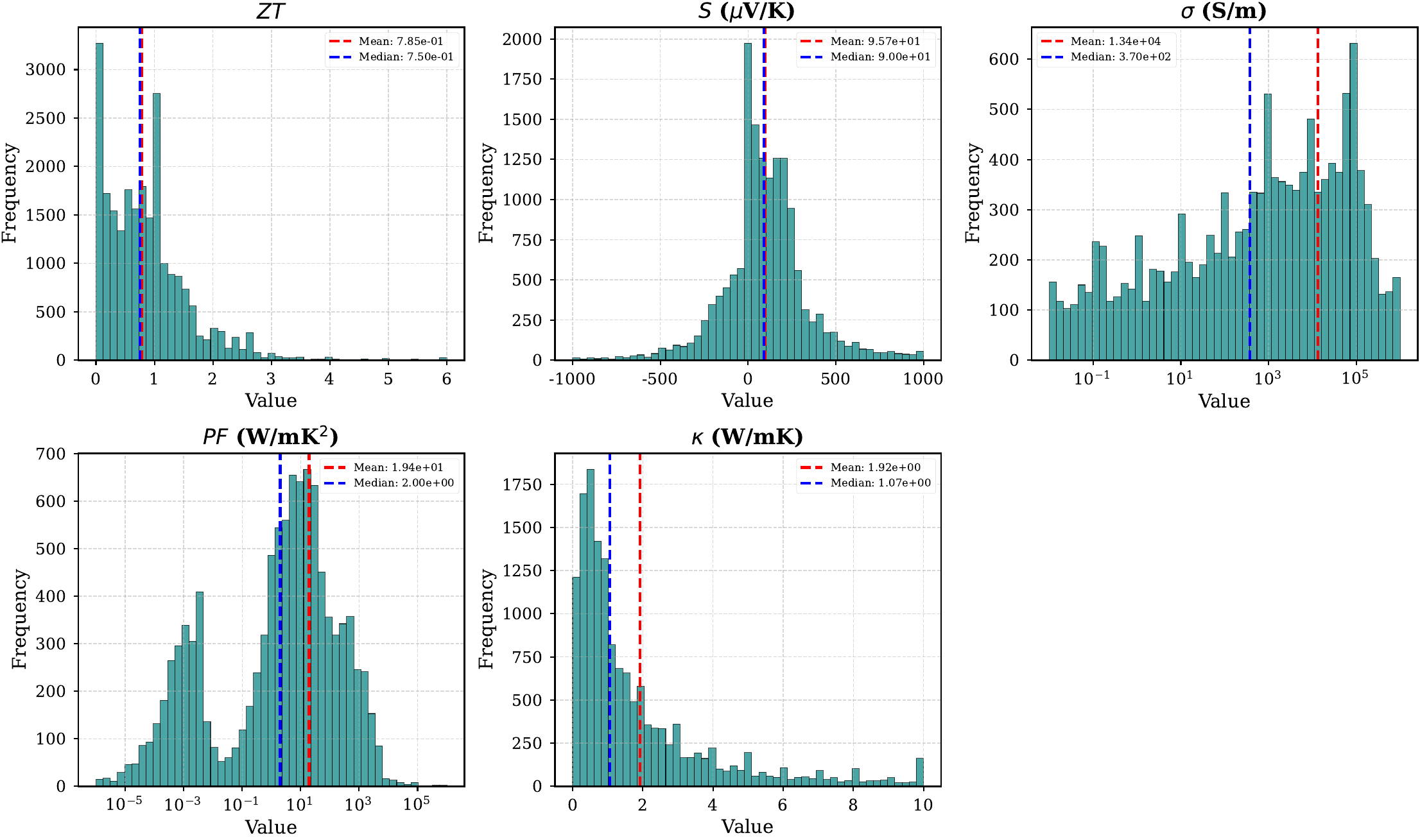}
    \caption{Distribution of normalized thermoelectric properties across the curated dataset. To avoid redundancy, electrical conductivity ($\sigma$) and resistivity ($\rho$) are merged into a single unified distribution. Vertical dashed lines indicate the mean (red) and median (blue) for each property.}
    \label{fig:distribution}
\end{figure}

To ensure consistency, the extracted property values from diverse literature sources were unit-normalized using basic NLP parsing and conversion rules. All Seebeck coefficients were expressed in $\mu$V/K, electrical conductivity in S/m, electrical resistivity in $\Omega\cdot$m, power factor in W/mK$^2$, and thermal conductivity in W/mK. Figure~\ref{fig:distribution} shows the resulting property distributions.  

We plotted $ZT$, Seebeck coefficient, and thermal conductivity on a linear scale since their ranges are relatively narrow and symmetric. In contrast, electrical conductivity and power factor span several orders of magnitude, making logarithmic scaling more suitable for capturing their spread. Electrical conductivity ($\sigma$) and resistivity ($\rho$), being inverse quantities, were combined into a single representation to avoid redundancy and ensure consistent statistics across the dataset.  

The summary statistics of each property are given below, with their distributions shown in Figure~\ref{fig:distribution}. The observed spread is strongly influenced by compound families, as different material classes exhibit characteristic transport behaviors. For example, chalcogenides typically contribute to the lower thermal conductivity regime \cite{li2020in4pb5,ojo2022structural,hobbis2020thermal,kurosaki2013low,osei2019ultralow}, while alloys and half-Heuslers extend the range of electrical conductivity and power factor \cite{khandy2021inspecting,ciesielski2020high,he2016achieving,mitra2022conventional}. Skutterudites and perovskites populate the mid-range $ZT$ distribution, whereas polymers and composites contribute to broader variability due to their structural heterogeneity.  

\begin{itemize}
    \item \textbf{Figure of merit ($ZT$)}: Mean = 0.785, Median = 0.750, Std.\ Dev. = 0.576.  
    The distribution is narrowly centered around the median, suggesting consistent reporting across studies. 
    Alloys, half-Heuslers, and skutterudites dominate the mid-range, with a long but sparse tail of higher-$ZT$ 
    entries reflecting optimized experimental systems.  

    \item \textbf{Seebeck Coefficient ($S$) ($\mu$V/K)}: Mean = 95.7, Median = 90.0, Std.\ Dev. = 171.3.  
    The histogram is tightly clustered around 0–200 $\mu$V/K, indicating a nearly symmetric profile. 
    Chalcogenides and oxides make up the bulk of entries, with occasional extreme values (positive and negative) 
    reflecting either unusual material classes or reporting artifacts.  

    \item \textbf{Electrical Conductivity ($\sigma$) (S/m)}: Mean = $1.34\times 10^4$, Median = 370, Std.\ Dev. = $2.81\times 10^4$.  
    The log-scale plot shows a broad right-skewed distribution: most compounds fall in the semiconducting regime 
    ($10^2$–$10^4$ S/m), while metallic alloys and heavily doped semiconductors drive the long high-conductivity tail.  

    \item \textbf{Power Factor (PF) (W/mK$^2$)}: Mean = 19.4, Median = 2.0, Std.\ Dev. = 42.0.  
    Most entries cluster close to the median, reflecting typical thermoelectric materials. 
    Outliers with exceptionally high PF are largely associated with optimized half-Heuslers and skutterudites, 
    consistent with their reputation for high-performance design.  

    \item \textbf{Thermal Conductivity ($\kappa$) (W/mK)}: Mean = 1.92, Median = 1.07, Std.\ Dev. = 2.16.  
    The majority of compounds lie in the 0.5–2 W/mK window, dominated by chalcogenides. 
    Oxides and perovskites extend the distribution toward higher $\kappa$, while a smaller subset of engineered 
    systems push values below 1 W/mK, desirable for high $ZT$.  
\end{itemize}

\subsection{Structural Properties}

\begin{figure}[H]
    \centering
    \includegraphics[width=0.85\textwidth]{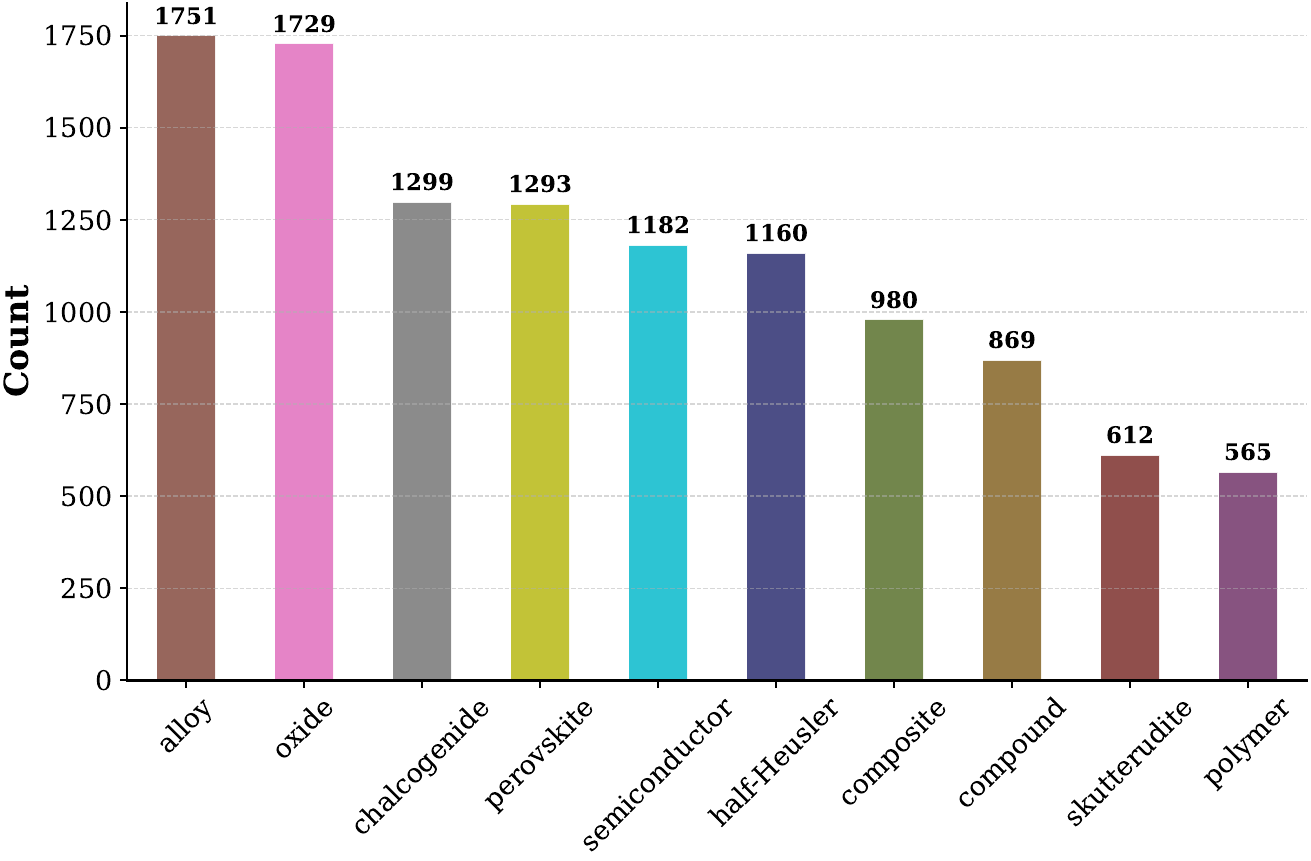}
    \caption{Top 10 compound types present in the dataset.}
    \label{fig:compound_type}
\end{figure}

\begin{figure}[H]
    \centering
    \includegraphics[width=0.85\textwidth]{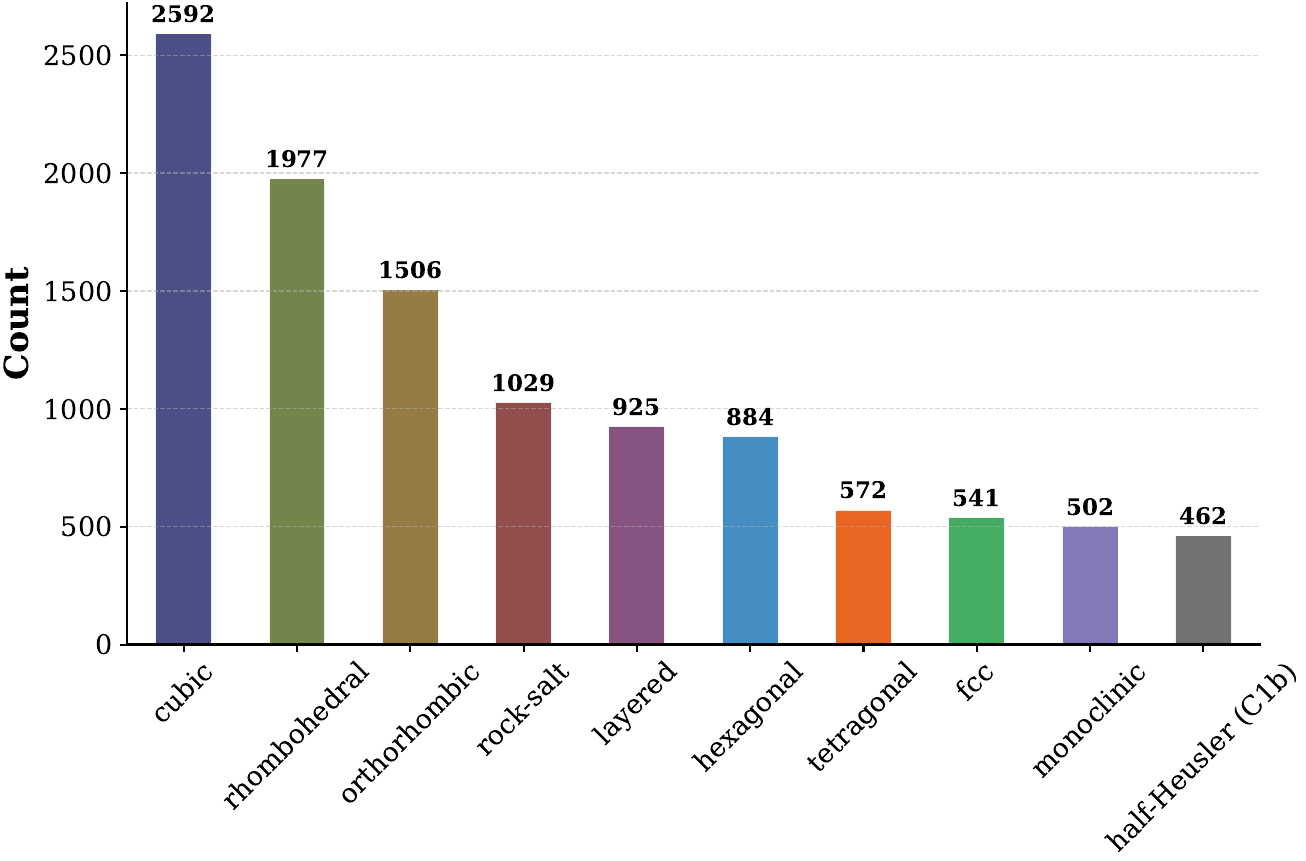}
    \caption{Top 10 crystal structures in the dataset.}
    \label{fig:crystal_structure}
\end{figure}

\begin{figure}[H]
    \centering
    \includegraphics[width=0.85\textwidth]{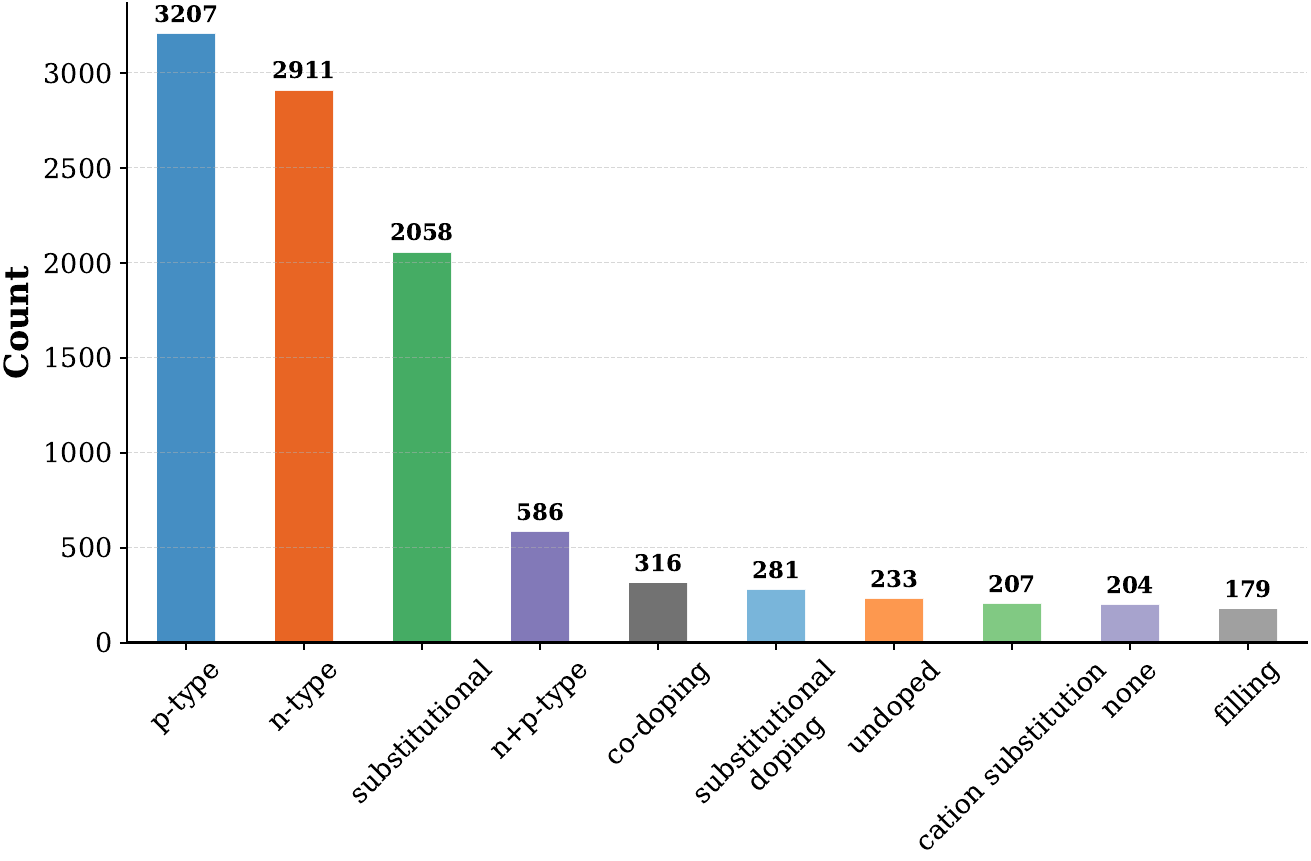}
    \caption{Top 10 doping types in the dataset.}
    \label{fig:doping_type}
\end{figure}

Alongside thermoelectric performance metrics, the dataset also captures structural descriptors that are critical for understanding structure–property correlations. Figures~\ref{fig:compound_type}--\ref{fig:doping_type} summarize the top ten categories for compound types, crystal structures, and doping strategies.  

Figure~\ref{fig:compound_type} shows that alloys, oxides, and chalcogenides dominate the dataset, followed by perovskites, semiconductors, and half-Heuslers. These classes are well-established in thermoelectric research: chalcogenides (e.g., tellurides, selenides) are known for their intrinsically low thermal conductivity, while oxides and perovskites are valued for stability and tunability. Skutterudites and half-Heuslers represent advanced crystalline families with proven potential for high $ZT$~\cite{shi2011multiple,quinn2021advances,chen2024thermoelectric}, whereas polymers and composites highlight growing interest in flexible and low-cost thermoelectrics~\cite{huo2022advances,he2017advances,wang2017polymer}.

In terms of crystal structure (Figure~\ref{fig:crystal_structure}), cubic symmetry is most prevalent, followed by rhombohedral and orthorhombic lattices. Rock-salt, layered, and hexagonal frameworks are also frequently observed, reflecting the structural motifs that enable favorable electronic transport and phonon scattering. Specific structure types, such as half-Heusler (C1b), appear as specialized subclasses, reflecting targeted materials engineering strategies.  

Doping types (Figure~\ref{fig:doping_type}) further emphasize the breadth of experimental strategies. The dataset contains a nearly balanced distribution of $p$-type (3207 entries) and $n$-type (2911 entries) materials, indicating broad exploration of both conduction polarities. Substitutional doping is the most widely employed modification method, with co-doping  appearing in smaller but significant numbers. The presence of both undoped and mixed ($n+p$-type) cases illustrates efforts to benchmark intrinsic behavior as well as explore band engineering for optimized performance.  

Together, these structural attributes complement the thermoelectric properties, enabling integrated analyses of how chemistry, symmetry, and doping govern transport phenomena across material families. It also allows downstream use of this dataset for predictive modeling structure-property correlations.

\subsection{Structure–Property Trends: ZT vs Temperature}

To further illustrate the impact of structural classes and doping on thermoelectric performance, we examined $ZT$ as a function of temperature for alloys and oxides.  

\begin{figure}[H]
    \centering
    \includegraphics[width=0.85\textwidth]{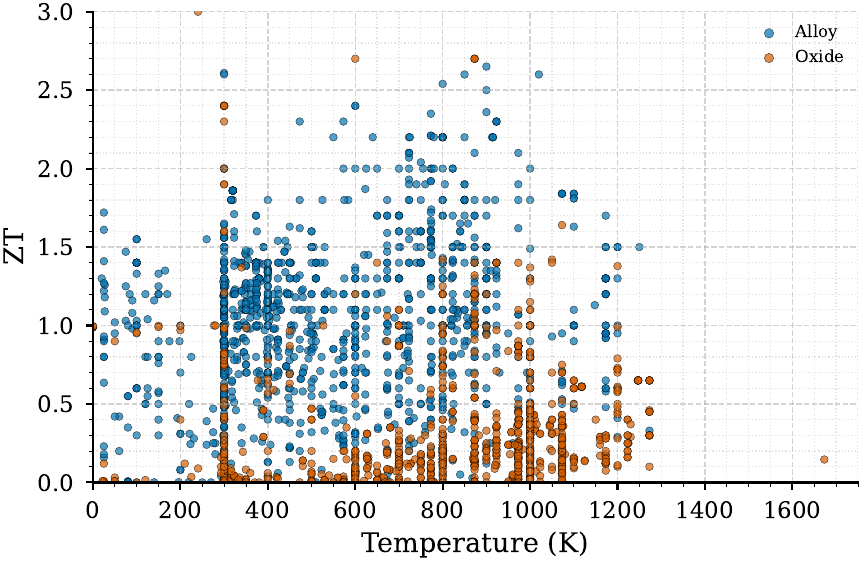}
    \caption{$ZT$ as a function of temperature for alloys and oxides. Alloys show a wider spread and higher maximum values compared to oxides.}
    \label{fig:ZT_alloy_oxide}
\end{figure}

Figure~\ref{fig:ZT_alloy_oxide} presents the raw $ZT$ vs temperature scatter for both families. Alloys exhibit a broad distribution with numerous entries above $ZT > 1$, whereas oxides are more concentrated at lower $ZT$, typically below unity. This contrast reflects the long-recognized difference between metallic alloys, which benefit from higher electrical conductivity, and oxides, which often suffer from relatively high thermal conductivity and lower carrier mobility.  

\begin{figure}[H]
    \centering
    \includegraphics[width=0.85\textwidth]{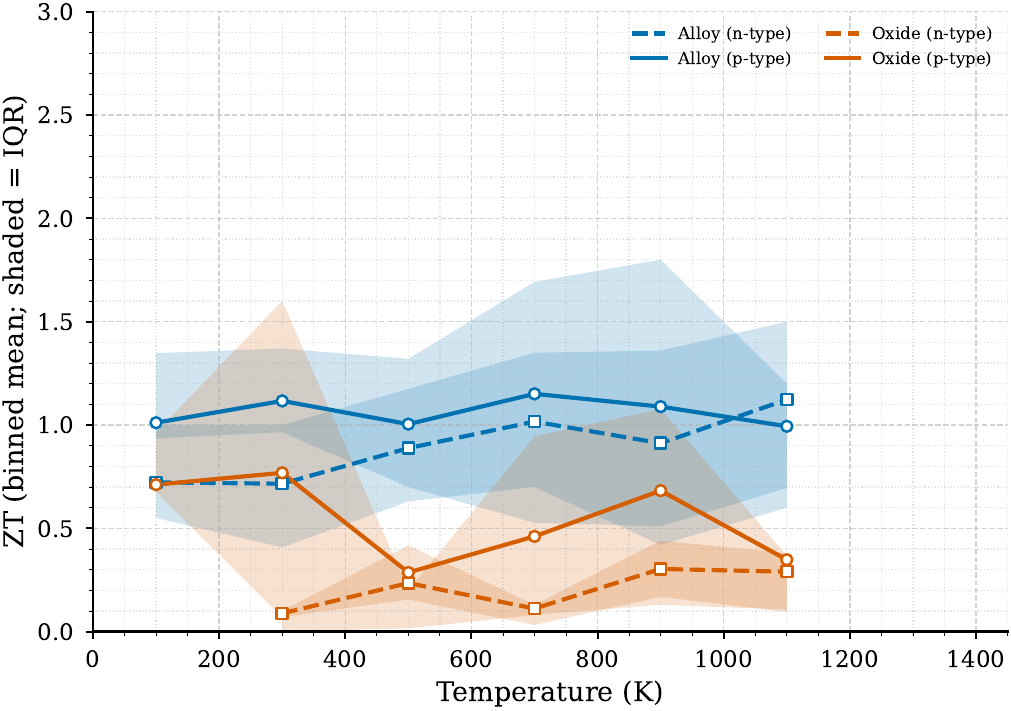}
    \caption{Binned $ZT$ vs temperature for alloys and oxides separated by doping type. Shaded regions denote the interquartile range (IQR). $p$-type alloys outperform other classes across the measured range.}
    \label{fig:ZT_doping}
\end{figure}

To disentangle doping effects, we further binned the data by doping type (Figure~\ref{fig:ZT_doping}). For alloys, $p$-type samples consistently outperform $n$-type across most temperature ranges, with median values near or above $ZT \sim 1$. Oxides, however, show limited performance overall, with both $p$- and $n$-type rarely exceeding $ZT \sim 0.8$. Notably, $p$-type alloys maintain stable $ZT$ over a wide thermal window, highlighting their potential for mid- to high-temperature thermoelectric applications.  

These observations confirm that both compound family and doping strategy strongly influence achievable thermoelectric performance. While alloys remain the most promising class in terms of high $ZT$, improving oxide-based thermoelectrics requires further engineering of electronic structure and phonon scattering.

\section{Interactive Dataset Explorer}

\begin{figure}[H]
    \centering
    \includegraphics[width=1.0\textwidth]{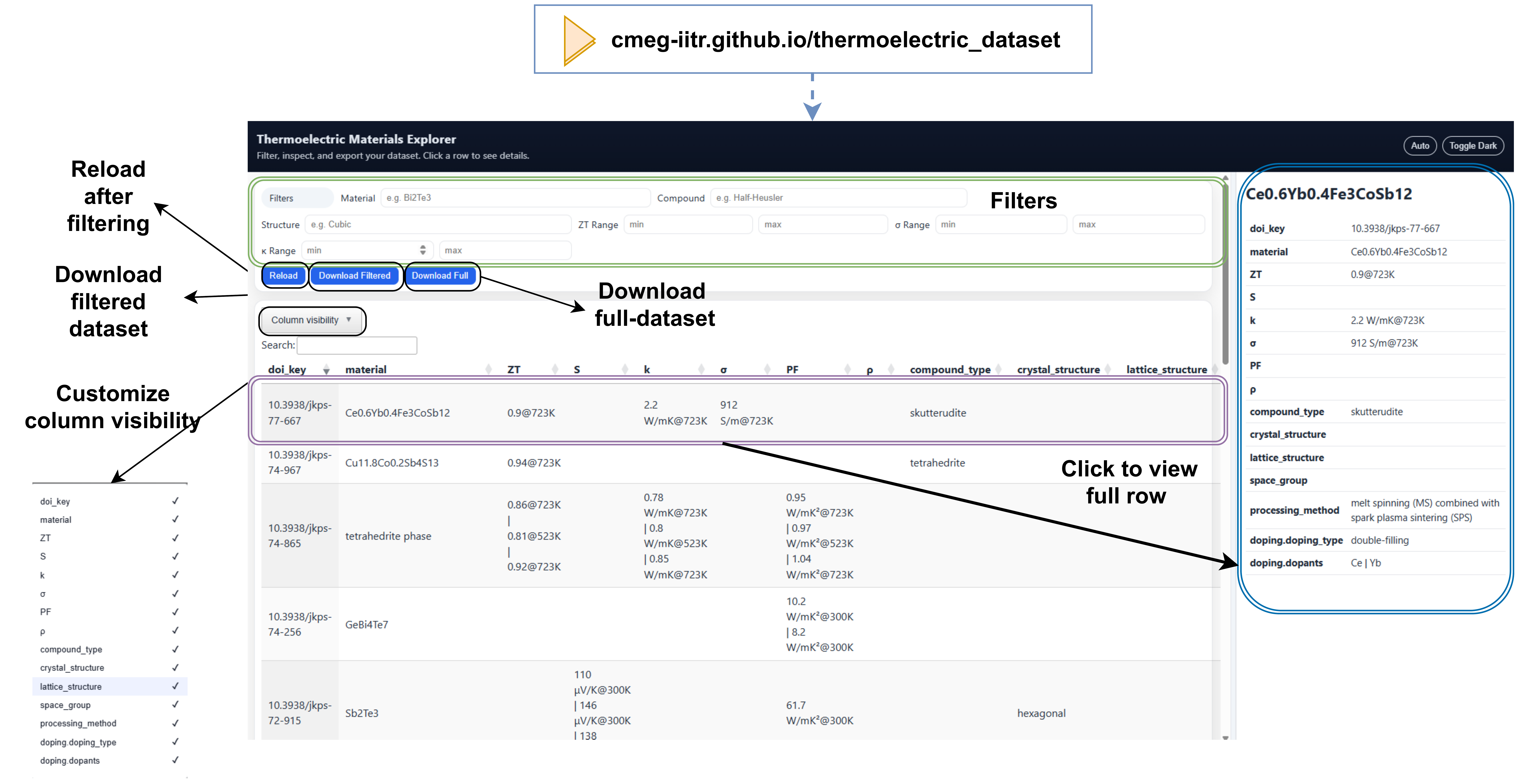}
    \caption{Thermoelectric Materials Explorer (\url{https://cmeg-iitr.github.io/thermoelectric_dataset}). Users can filter by semantic attributes and numeric property ranges, inspect rows with a details pane, toggle column visibility, and export filtered results.}
    \label{fig:explorer}
\end{figure}

To facilitate inspection and reuse, we developed a lightweight web interface\footnote{\url{https://cmeg-iitr.github.io/thermoelectric_dataset}} (Figure~\ref{fig:explorer}) for the curated dataset comprising \textbf{27{,}822} entries. The tool enables rapid query, visual triage, and export without requiring local setup.

\begin{itemize}
    \item \textbf{Semantic filters:} search by \textit{material name}, \textit{compound type}, and \textit{crystal structure}.
    \item \textbf{Numeric range filters:} bounded sliders/inputs for key TE properties ($ZT$, $\sigma$, $\kappa$) to isolate regimes of interest.
    \item \textbf{Dynamic table view:} interactive grid with \textit{column-visibility} controls to toggle attributes on/off.
    \item \textbf{Details pane:} on row selection, the right panel displays full metadata and extracted properties (including temperatures, processing method, doping type/dopants).
    \item \textbf{Export:} one-click download of \textit{filtered subsets} or the \textit{entire dataset} in CSV format to support downstream analysis.
\end{itemize}

This explorer serves as a reproducible front end for hypothesis generation (e.g., filtering by structure class and thermal-conductivity window) and for assembling task-specific benchmarks.

\section{Conclusion}
In this work, we present a large-scale, agentic LLM-based workflow for automated data extraction from the scientific literature. By combining modular agents, dynamic token allocation, conditional table parsing, and rigorous benchmarking across multiple state-of-the-art LLMs, our framework demonstrates both high accuracy and scalability. Applied to $\sim$10,000 full-text articles, the pipeline curated 27,822 temperature-resolved records spanning $ZT$, Seebeck coefficient, electrical conductivity/resistivity, power factor, thermal conductivity, and structural attributes such as crystal class, space group, and doping strategy, which is one of the largest structure--property datasets of thermoelectric materials. The resulting corpus not only reproduces established thermoelectric trends such as the superior performance of alloys over oxides and the advantage of $p$-type doping but also exposes broader structure--property correlations that are challenging to capture in traditional databases.

Beyond the dataset itself, we emphasize the importance of transparent cost--quality trade-offs in LLM-driven data mining. Our results show that GPT-4.1 achieves the highest accuracy ($F1 \approx 0.91$), while GPT-4.1-mini offers nearly comparable performance ($F1 \approx 0.89$) at substantially reduced computational cost, enabling practical deployment at scale. The curated dataset is made accessible through an interactive web explorer that supports semantic queries, property-range filtering, and export for downstream machine learning tasks, thereby lowering barriers for adoption within the community.

Taken together, this study establishes a reproducible, cost profiled, and extensible paradigm for curating structure–property datasets directly from unstructured literature. While we have focused here on thermoelectrics, the modular agent design and zero-shot adaptability render the workflow broadly generalizable to other functional materials domains including batteries, catalysts, and magnetic materials by adjusting schema and prompt templates. We anticipate that this approach will accelerate hypothesis generation, guide machine-learning model development, and ultimately contribute to shortening the discovery cycle for next-generation materials.
\section*{Code Availability}

The code used for data extraction and analysis in this study is available at:

\medskip

\noindent
\url{https://github.com/CMEG-IITR/Agentic_data_extraction}

\medskip
 
\bibliographystyle{unsrt}
\bibliography{references}

\begin{thebibliography}{10}

\bibitem{trewartha2022quantifying}
Amalie Trewartha, Nicholas Walker, Haoyan Huo, Sanghoon Lee, Kevin Cruse, John Dagdelen, Alexander Dunn, Kristin~A Persson, Gerbrand Ceder, and Anubhav Jain.
\newblock Quantifying the advantage of domain-specific pre-training on named entity recognition tasks in materials science.
\newblock {\em Patterns}, 3(4), 2022.

\bibitem{chithrananda2020chemberta}
Seyone Chithrananda, Gabriel Grand, and Bharath Ramsundar.
\newblock Chemberta: large-scale self-supervised pretraining for molecular property prediction.
\newblock {\em arXiv preprint arXiv:2010.09885}, 2020.

\bibitem{shetty2023general}
Pranav Shetty, Arunkumar~Chitteth Rajan, Chris Kuenneth, Sonakshi Gupta, Lakshmi~Prerana Panchumarti, Lauren Holm, Chao Zhang, and Rampi Ramprasad.
\newblock A general-purpose material property data extraction pipeline from large polymer corpora using natural language processing.
\newblock {\em npj Computational Materials}, 9(1):52, 2023.

\bibitem{dagdelen2024structured}
John Dagdelen, Alexander Dunn, Sanghoon Lee, Nicholas Walker, Andrew~S Rosen, Gerbrand Ceder, Kristin~A Persson, and Anubhav Jain.
\newblock Structured information extraction from scientific text with large language models.
\newblock {\em Nature communications}, 15(1):1418, 2024.

\bibitem{zheng2023chatgpt}
Zhiling Zheng, Oufan Zhang, Christian Borgs, Jennifer~T Chayes, and Omar~M Yaghi.
\newblock Chatgpt chemistry assistant for text mining and the prediction of mof synthesis.
\newblock {\em Journal of the American Chemical Society}, 145(32):18048--18062, 2023.

\bibitem{polak2024extracting}
Maciej~P Polak and Dane Morgan.
\newblock Extracting accurate materials data from research papers with conversational language models and prompt engineering.
\newblock {\em Nature Communications}, 15(1):1569, 2024.

\bibitem{yang2023accurate}
Samuel~J Yang, Shutong Li, Subhashini Venugopalan, Vahe Tshitoyan, Muratahan Aykol, Amil Merchant, Ekin~Dogus Cubuk, and Gowoon Cheon.
\newblock Accurate prediction of experimental band gaps from large language model-based data extraction.
\newblock {\em arXiv preprint arXiv:2311.13778}, 2023.

\bibitem{gupta2024data}
Sonakshi Gupta, Akhlak Mahmood, Pranav Shetty, Aishat Adeboye, and Rampi Ramprasad.
\newblock Data extraction from polymer literature using large language models.
\newblock {\em Communications materials}, 5(1):269, 2024.

\bibitem{Ansari2024}
M.~Ansari and S.~M. Moosavi.
\newblock Agent-based learning of materials datasets from the scientific literature.
\newblock {\em Digital Discovery}, 3(12):2607--2617, 2024.

\bibitem{li2022machine}
Man Li, Lingyun Dai, and Yongjie Hu.
\newblock Machine learning for harnessing thermal energy: From materials discovery to system optimization.
\newblock {\em ACS energy letters}, 7(10):3204--3226, 2022.

\bibitem{jain2016computational}
Anubhav Jain, Yongwoo Shin, and Kristin~A Persson.
\newblock Computational predictions of energy materials using density functional theory.
\newblock {\em Nature Reviews Materials}, 1(1):1--13, 2016.

\bibitem{gorai2017computationally}
Prashun Gorai, Vladan Stevanovi{\'c}, and Eric~S Toberer.
\newblock Computationally guided discovery of thermoelectric materials.
\newblock {\em Nature Reviews Materials}, 2(9):1--16, 2017.

\bibitem{deng2024high}
Tingting Deng, Pengfei Qiu, Tingwei Yin, Ze~Li, Jiong Yang, Tianran Wei, and Xun Shi.
\newblock High-throughput strategies in the discovery of thermoelectric materials.
\newblock {\em Advanced Materials}, 36(13):2311278, 2024.

\bibitem{sarikurt2020high}
Sevil Sarikurt, Tu{\u{g}}bey Kocaba{\c{s}}, and Cem Sevik.
\newblock High-throughput computational screening of 2d materials for thermoelectrics.
\newblock {\em Journal of Materials Chemistry A}, 8(37):19674--19683, 2020.

\bibitem{chelikowsky2011computational}
James~R Chelikowsky, MMG Alemany, TL~Chan, and GM~Dalpian.
\newblock Computational studies of doped nanostructures.
\newblock {\em Reports on Progress in Physics}, 74(4):046501, 2011.

\bibitem{jia2022unsupervised}
Xue Jia, Yanshuai Deng, Xin Bao, Honghao Yao, Shan Li, Zhou Li, Chen Chen, Xinyu Wang, Jun Mao, Feng Cao, et~al.
\newblock Unsupervised machine learning for discovery of promising half-heusler thermoelectric materials.
\newblock {\em npj Computational Materials}, 8(1):34, 2022.

\bibitem{wang2020machine}
Tian Wang, Cheng Zhang, Hichem Snoussi, and Gang Zhang.
\newblock Machine learning approaches for thermoelectric materials research.
\newblock {\em Advanced Functional Materials}, 30(5):1906041, 2020.

\bibitem{wang2023critical}
Xiangdong Wang, Ye~Sheng, Jinyan Ning, Jinyang Xi, Lili Xi, Di~Qiu, Jiong Yang, and Xuezhi Ke.
\newblock A critical review of machine learning techniques on thermoelectric materials.
\newblock {\em The Journal of Physical Chemistry Letters}, 14(7):1808--1822, 2023.

\bibitem{antunes2023predicting}
Luis~M Antunes, Keith~T Butler, and Ricardo Grau-Crespo.
\newblock Predicting thermoelectric transport properties from composition with attention-based deep learning.
\newblock {\em Machine Learning: Science and Technology}, 4(1):015037, 2023.

\bibitem{sparks2016data}
Taylor~D Sparks, Michael~W Gaultois, Anton Oliynyk, Jakoah Brgoch, and Bryce Meredig.
\newblock Data mining our way to the next generation of thermoelectrics.
\newblock {\em Scripta Materialia}, 111:10--15, 2016.

\bibitem{mbaye2021data}
Mamadou~T Mbaye, Sangram~K Pradhan, and Messaoud Bahoura.
\newblock Data-driven thermoelectric modeling: Current challenges and prospects.
\newblock {\em Journal of Applied Physics}, 130(19), 2021.

\bibitem{ricci2017ab}
Francesco Ricci, Wei Chen, Umut Aydemir, G~Jeffrey Snyder, Gian-Marco Rignanese, Anubhav Jain, and Geoffroy Hautier.
\newblock An ab initio electronic transport database for inorganic materials.
\newblock {\em Scientific data}, 4(1):1--13, 2017.

\bibitem{choudhary2020joint}
Kamal Choudhary, Kevin~F Garrity, Andrew~CE Reid, Brian DeCost, Adam~J Biacchi, Angela~R Hight~Walker, Zachary Trautt, Jason Hattrick-Simpers, A~Gilad Kusne, Andrea Centrone, et~al.
\newblock The joint automated repository for various integrated simulations (jarvis) for data-driven materials design.
\newblock {\em npj computational materials}, 6(1):173, 2020.

\bibitem{yao2021materials}
Mingjia Yao, Yuxiang Wang, Xin Li, Ye~Sheng, Haiyang Huo, Lili Xi, Jiong Yang, and Wenqing Zhang.
\newblock Materials informatics platform with three dimensional structures, workflow and thermoelectric applications.
\newblock {\em Scientific Data}, 8(1):236, 2021.

\bibitem{gorai2016te}
Prashun Gorai, Duanfeng Gao, Brenden Ortiz, Sam Miller, Scott~A Barnett, Thomas Mason, Qin Lv, Vladan Stevanovi{\'c}, and Eric~S Toberer.
\newblock Te design lab: A virtual laboratory for thermoelectric material design.
\newblock {\em Computational Materials Science}, 112:368--376, 2016.

\bibitem{wang2011assessing}
Shidong Wang, Zhao Wang, Wahyu Setyawan, Natalio Mingo, and Stefano Curtarolo.
\newblock Assessing the thermoelectric properties of sintered compounds<? format?> via high-throughput ab-initio calculations.
\newblock {\em Physical Review X}, 1(2):021012, 2011.

\bibitem{carrete2014finding}
Jes{\'u}s Carrete, Wu~Li, Natalio Mingo, Shidong Wang, and Stefano Curtarolo.
\newblock Finding unprecedentedly low-thermal-conductivity half-heusler semiconductors via high-throughput materials modeling.
\newblock {\em Physical Review X}, 4(1):011019, 2014.

\bibitem{xi2018discovery}
Lili Xi, Shanshan Pan, Xin Li, Yonglin Xu, Jianyue Ni, Xin Sun, Jiong Yang, Jun Luo, Jinyang Xi, Wenhao Zhu, et~al.
\newblock Discovery of high-performance thermoelectric chalcogenides through reliable high-throughput material screening.
\newblock {\em Journal of the American Chemical Society}, 140(34):10785--10793, 2018.

\bibitem{fang2024wenzhou}
Ying Fang and Hezhu Shao.
\newblock Wenzhou te: A first-principle-calculated thermoelectric materials database.
\newblock {\em Materials}, 17(10):2200, 2024.

\bibitem{na2022public}
Gyoung~S Na and Hyunju Chang.
\newblock A public database of thermoelectric materials and system-identified material representation for data-driven discovery.
\newblock {\em npj Computational Materials}, 8(1):214, 2022.

\bibitem{katsura2019data}
Yukari Katsura, Masaya Kumagai, Takushi Kodani, Mitsunori Kaneshige, Yuki Ando, Sakiko Gunji, Yoji Imai, Hideyasu Ouchi, Kazuki Tobita, Kaoru Kimura, et~al.
\newblock Data-driven analysis of electron relaxation times in pbte-type thermoelectric materials.
\newblock {\em Science and Technology of Advanced Materials}, 20(1):511--520, 2019.

\bibitem{gaultois2013data}
Michael~W Gaultois, Taylor~D Sparks, Christopher~KH Borg, Ram Seshadri, William~D Bonificio, and David~R Clarke.
\newblock Data-driven review of thermoelectric materials: performance and resource considerations.
\newblock {\em Chemistry of Materials}, 25(15):2911--2920, 2013.

\bibitem{priya2021accelerated}
Pikee Priya and Narayana~R Aluru.
\newblock Accelerated design and discovery of perovskites with high conductivity for energy applications through machine learning.
\newblock {\em npj Computational Materials}, 7(1):90, 2021.

\bibitem{lee2023texplorer}
Yea-Lee Lee, Hyungseok Lee, Seunghun Jang, Jeongho Shin, Taeshik Kim, Sejin Byun, In~Chung, Jino Im, and Hyunju Chang.
\newblock Texplorer. org: Thermoelectric material properties data platform for experimental and first-principles calculation results.
\newblock {\em APL Materials}, 11(4), 2023.

\bibitem{mavracic2021chemdataextractor}
Juraj Mavracic, Callum~J Court, Taketomo Isazawa, Stephen~R Elliott, and Jacqueline~M Cole.
\newblock Chemdataextractor 2.0: Autopopulated ontologies for materials science.
\newblock {\em Journal of Chemical Information and Modeling}, 61(9):4280--4289, 2021.

\bibitem{sierepeklis2022thermoelectric}
Odysseas Sierepeklis and Jacqueline~M Cole.
\newblock A thermoelectric materials database auto-generated from the scientific literature using chemdataextractor.
\newblock {\em Scientific Data}, 9(1):648, 2022.

\bibitem{itani2025large}
Suman Itani, Yibo Zhang, and Jiadong Zang.
\newblock Large language model-driven database for thermoelectric materials.
\newblock {\em Computational Materials Science}, 253:113855, 2025.

\bibitem{zhang2024gptarticleextractor}
Yibo Zhang, Suman Itani, Kamal Khanal, Emmanuel Okyere, Gavin Smith, Koichiro Takahashi, and Jiadong Zang.
\newblock Gptarticleextractor: An automated workflow for magnetic material database construction.
\newblock {\em Journal of Magnetism and Magnetic Materials}, 597:172001, 2024.

\bibitem{olivetti2020data}
Elsa~A Olivetti, Jacqueline~M Cole, Edward Kim, Olga Kononova, Gerbrand Ceder, Thomas Yong-Jin Han, and Anna~M Hiszpanski.
\newblock Data-driven materials research enabled by natural language processing and information extraction.
\newblock {\em Applied Physics Reviews}, 7(4), 2020.

\bibitem{smith2022challenges}
Andrew Smith, Vinayak Bhat, Qianxiang Ai, and Chad Risko.
\newblock Challenges in information-mining the materials literature: a case study and perspective.
\newblock {\em Chemistry of Materials}, 34(11):4821--4827, 2022.

\bibitem{blecher2023nougat}
Lukas Blecher, Guillem Cucurull, Thomas Scialom, and Robert Stojnic.
\newblock Nougat: Neural optical understanding for academic documents.
\newblock {\em arXiv preprint arXiv:2308.13418}, 2023.

\bibitem{paruchuri2023marker}
V.~Paruchuri.
\newblock Marker: Open source machine learning model for data annotation.
\newblock \url{https://github.com/VikParuchuri/marker}, 2023.

\bibitem{oka2021machine}
Hiroyuki Oka, Atsushi Yoshizawa, Hiroyuki Shindo, Yuji Matsumoto, and Masashi Ishii.
\newblock Machine extraction of polymer data from tables using xml versions of scientific articles.
\newblock {\em Science and Technology of Advanced Materials: Methods}, 1(1):12--23, 2021.

\bibitem{chatgpt2025}
OpenAI.
\newblock Chatgpt.
\newblock \url{https://chat.openai.com/}, 2025.
\newblock Large language model accessed for generating regular expressions from keywords.

\bibitem{tiktoken}
{OpenAI}.
\newblock tiktoken.
\newblock \url{https://github.com/openai/tiktoken}, 2022.
\newblock Available at: \url{https://github.com/openai/tiktoken}.

\bibitem{langgraph2024}
LangChain AI.
\newblock Langgraph.
\newblock \url{https://github.com/langchain-ai/langgraph}, 2024.

\bibitem{renze2024effect}
Matthew Renze.
\newblock The effect of sampling temperature on problem solving in large language models.
\newblock In {\em Findings of the association for computational linguistics: EMNLP 2024}, pages 7346--7356, 2024.

\bibitem{zhu2024hot}
Yuqi Zhu, Jia Li, Ge~Li, YunFei Zhao, Jia Li, Zhi Jin, and Hong Mei.
\newblock Hot or cold? adaptive temperature sampling for code generation with large language models.
\newblock In {\em Proceedings of the Thirty-Eighth AAAI Conference on Artificial Intelligence and Thirty-Sixth Conference on Innovative Applications of Artificial Intelligence and Fourteenth Symposium on Educational Advances in Artificial Intelligence}, AAAI'24/IAAI'24/EAAI'24. AAAI Press, 2024.

\bibitem{liu2023lost}
Nelson~F Liu, Kevin Lin, John Hewitt, Ashwin Paranjape, Michele Bevilacqua, Fabio Petroni, and Percy Liang.
\newblock Lost in the middle: How language models use long contexts.
\newblock {\em arXiv preprint arXiv:2307.03172}, 2023.

\bibitem{openai_pricing}
{OpenAI}.
\newblock Openai api pricing.
\newblock \url{https://openai.com/api/pricing/}, 2025.

\bibitem{google_gemini_pricing}
{Google DeepMind}.
\newblock Gemini api pricing.
\newblock \url{https://ai.google.dev/gemini-api/docs/pricing}, 2025.
\newblock Accessed: 2025-08-10.

\bibitem{li2020in4pb5}
Jingpeng Li, Shiqiang Hao, Shangqing Qu, Christopher Wolverton, Jing Zhao, and Yonggang Wang.
\newblock In4pb5. 5sb5s19: A stable quaternary chalcogenide with low thermal conductivity.
\newblock {\em Inorganic chemistry}, 60(1):325--333, 2020.

\bibitem{ojo2022structural}
Oluwagbemiga~P Ojo, Wilarachchige~DCB Gunatilleke, Hsin Wang, and George~S Nolas.
\newblock Structural and thermal properties of ultralow thermal conductivity ba 3 cu 2 sn 3 se 10.
\newblock {\em Dalton Transactions}, 51(16):6220--6225, 2022.

\bibitem{hobbis2020thermal}
Dean Hobbis, Hsin Wang, Joshua Martin, and George~S Nolas.
\newblock Thermal properties of the very low thermal conductivity ternary chalcogenide cu4bi4m9 (m= s, se).
\newblock {\em physica status solidi (RRL)--Rapid Research Letters}, 14(8):2000166, 2020.

\bibitem{kurosaki2013low}
Ken Kurosaki and Shinsuke Yamanaka.
\newblock Low-thermal-conductivity group 13 chalcogenides as high-efficiency thermoelectric materials, 2013.

\bibitem{osei2019ultralow}
Eric Osei-Agyemang, Challen~Enninful Adu, and Ganesh Balasubramanian.
\newblock Ultralow lattice thermal conductivity of chalcogenide perovskite cazrse3 contributes to high thermoelectric figure of merit.
\newblock {\em npj Computational Materials}, 5(1):116, 2019.

\bibitem{khandy2021inspecting}
Shakeel~Ahmad Khandy.
\newblock Inspecting the electronic structure and thermoelectric power factor of novel p-type half-heuslers.
\newblock {\em Scientific reports}, 11(1):20756, 2021.

\bibitem{ciesielski2020high}
Kamil Ciesielski, Karol Synoradzki, I~Wola{\'n}ska, Piotr Stachowiak, L~K{\c{e}}pi{\'n}ski, Andrzej Je{\.z}owski, Tomasz Toli{\'n}ski, and Dariusz Kaczorowski.
\newblock High-temperature power factor of half-heusler phases renisb (re = sc, dy, ho, er, tm, lu).
\newblock {\em Journal of Alloys and Compounds}, 816:152596, 2020.

\bibitem{he2016achieving}
Ran He, Daniel Kraemer, Jun Mao, Lingping Zeng, Qing Jie, Yucheng Lan, Chunhua Li, Jing Shuai, Hee~Seok Kim, Yuan Liu, et~al.
\newblock Achieving high power factor and output power density in p-type half-heuslers nb1-xtixfesb.
\newblock {\em Proceedings of the National Academy of Sciences}, 113(48):13576--13581, 2016.

\bibitem{mitra2022conventional}
Mousumi Mitra, Allen Benton, Md~Sabbir Akhanda, Jie Qi, Mona Zebarjadi, David~J Singh, and S~Joseph Poon.
\newblock Conventional half-heusler alloys advance state-of-the-art thermoelectric properties.
\newblock {\em Materials Today Physics}, 28:100900, 2022.

\bibitem{shi2011multiple}
Xun Shi, Jiong Yang, James~R Salvador, Miaofang Chi, Jung~Y Cho, Hsin Wang, Shengqiang Bai, Jihui Yang, Wenqing Zhang, and Lidong Chen.
\newblock Multiple-filled skutterudites: high thermoelectric figure of merit through separately optimizing electrical and thermal transports.
\newblock {\em Journal of the American Chemical Society}, 133(20):7837--7846, 2011.

\bibitem{quinn2021advances}
Robert~J Quinn and Jan-Willem~G Bos.
\newblock Advances in half-heusler alloys for thermoelectric power generation.
\newblock {\em Materials Advances}, 2(19):6246--6266, 2021.

\bibitem{chen2024thermoelectric}
Rongchun Chen, Huijun Kang, Ruonan Min, Zongning Chen, Enyu Guo, Xiong Yang, and Tongmin Wang.
\newblock Thermoelectric properties of half-heusler alloys.
\newblock {\em International Materials Reviews}, 69(2):83--106, 2024.

\bibitem{huo2022advances}
Bingchen Huo and Cun-Yue Guo.
\newblock Advances in thermoelectric composites consisting of conductive polymers and fillers with different architectures.
\newblock {\em Molecules}, 27(20):6932, 2022.

\bibitem{he2017advances}
Jian He and Terry~M Tritt.
\newblock Advances in thermoelectric materials research: Looking back and moving forward.
\newblock {\em Science}, 357(6358):eaak9997, 2017.

\bibitem{wang2017polymer}
Liming Wang, Yuchen Liu, Zimeng Zhang, Biran Wang, Jingjing Qiu, David Hui, and Shiren Wang.
\newblock Polymer composites-based thermoelectric materials and devices.
\newblock {\em Composites Part B: Engineering}, 122:145--155, 2017.

\end{thebibliography}

\end{document}